\documentclass[sigconf]{acmart}

\usepackage{tikz}
\usetikzlibrary{arrows.meta,calc,positioning,shapes.geometric}

\definecolor{diagramnavy}{HTML}{17365D}
\definecolor{diagramblue}{HTML}{DCEAF7}
\definecolor{diagramgray}{HTML}{F3F5F7}
\definecolor{diagramteal}{HTML}{247B7B}
\definecolor{diagramteallight}{HTML}{E2F1F0}
\definecolor{diagramgreen}{HTML}{3A7D44}
\definecolor{diagramgreenlight}{HTML}{E5F2E7}
\definecolor{diagramgold}{HTML}{A66A00}
\definecolor{diagramgoldlight}{HTML}{FAEBC8}

\AtBeginDocument{%
  }

\setcopyright{rightsretained}
\copyrightyear{2026}
\acmYear{2026}
\acmDOI{}
\acmConference[GRAIL @ CIKM '26]%
{GRAIL 2026: Workshop on Generative, Retrieval-augmented, and Agentic Intelligence for Personnalization}%
{November 8, 2026}{Rome, Italy}
\acmISBN{}

\begin{document}

\title{A Technique to Model Product Information and User Reviews (MPIUR)}


\author{Emna Belguith}
\email{emna.belghith@supcom.tn}
\authornote{The first author's home institution is the Higher School of Communication of Tunis, but this work was conducted during her graduation internship at LTCI.}
\affiliation{%
  \institution{Department of Computer Science and Networks}
  \institution{Higher School of Communication of Tunis}
  \city{Tunis}
  \country{Tunisia}
}
\affiliation{%
  \institution{LTCI, T\'el\'ecom Paris, IP Paris}
  \city{Palaiseau}
  \country{France}
}

\author{Fajrian Yunus}
\email{fajrian.yunus@telecom-paris.fr}
\affiliation{%
  \institution{LTCI, T\'el\'ecom Paris, IP Paris}
  \city{Palaiseau}
  \country{France}
}

\author{Talel Abdessalem}
\email{talel.abdessalem@telecom-paris.fr}
\affiliation{%
  \institution{LTCI, T\'el\'ecom Paris, IP Paris}
  \city{Palaiseau}
  \country{France}
}

\renewcommand{\shortauthors}{Belguith, Yunus \& Abdessalem}

\begin{abstract}
Modern e-commerce platforms naturally provide three complementary modalities: textual item descriptions, item images, and user reviews. However, existing multimodal recommendation methods often omit reviews or do not integrate all three modalities within a unified framework. In this paper, we propose MPIUR, a multimodal graph-based recommendation model that encodes reviews in heterogeneous graphs, aligns image and text representations, and adapts collaborative--semantic alignment to item interaction density. Experiments on the Amazon Reviews 2023 Musical Instruments and Baby categories show that MPIUR achieves the best global nDCG@10 and Recall@10 among the evaluated methods on both datasets. It also obtains the best nDCG@10 and Recall@10 on sparse items, indicating that jointly exploiting reviews and multimodal item content is particularly useful when interaction histories are limited.
\end{abstract}




\maketitle

\section{Introduction}
Recommender systems rely heavily on user-item interaction data to capture user preferences. Although collaborative filtering has proven effective for popular items, its performance degrades when interaction data is sparse --- a pervasive issue in e-commerce, where many products have limited interaction histories. To compensate for this weakness, recent work has integrated multimodal item content, such as textual descriptions and images. However, existing methods either ignore user reviews or fail to integrate all three naturally available modalities — user reviews, textual item descriptions, and item images — in a unified framework.

Combining collaborative and semantic signals introduces two fundamental challenges. First, the \textit{collaborative-semantic tradeoff}: collaborative signals capture behavioral patterns from user-item interactions and excel at modeling popular items with rich interaction histories. However, for items with few interactions, these signals become unreliable and fail to produce meaningful representations. Conversely, semantic signals derived from multimodal item content can remain informative regardless of interaction volume, but relying on them excessively can obscure behavioral differences between users or items that look similar or have similar descriptions but do not necessarily attract the same users. Existing methods either prioritize one signal over the other or apply a static fusion strategy, failing to adapt to the varying interaction density of each item.

The \textit{modality gap} is another problem: even when both collaborative and semantic representations are available, they are learned in separate optimization processes and converge to geometrically distinct regions of the embedding space. As a result, naively combining them produces inconsistent representations — the collaborative space encodes behavioral proximity while the semantic space encodes content similarity, and the two do not naturally align. This limits the model's ability to effectively leverage both signals simultaneously.

In this paper, we propose a multimodal recommendation framework that addresses both challenges simultaneously. We construct heterogeneous graphs containing users, items, aspects, and review triplets; use review embeddings as features of user--item interactions; and fuse item images and textual descriptions by per-item attention. A popularity-weighted alignment loss places a stronger collaborative--multimodal constraint on sparse items, while the remaining objectives are applied uniformly.

The main contributions of this work are as follows:

\begin{itemize}

\item A unified framework jointly exploiting user reviews, textual item descriptions, and item images within a heterogeneous graph structure, where reviews serve as edge features enabling fine-grained interaction modeling.

\item A Modal Encoder that fuses item visual and textual representations via per-item adaptive attention and contrastive alignment.

\item An adaptive alignment objective that gives greater weight to multimodal information for items with fewer interactions.

\item Experiments on two Amazon datasets demonstrating the strongest and most consistent gains on less popular items, together with competitive results on popular items.

\end{itemize}

\section{Related Work}

\subsection{Recommender Systems}

Two classical paradigms underpin recommender systems: collaborative filtering and content-based filtering.

Collaborative filtering is based on the intuition that users who have interacted with similar items in the past likely have similar preferences. A system can therefore recommend items previously selected by similar users. Interactions may take various forms, including purchases, clicks, ratings, and reviews. In this work, we focus on interactions associated with reviews, which provide richer semantic information than clicks or purchases alone. A classical collaborative-filtering technique is matrix factorization~\cite{koren2009matrix}. More recent approaches use graph neural networks (GNNs), representing users and items as nodes and interactions as edges~\cite{wei2019mmgcn, wang2021dualgnn}. The main limitation of collaborative filtering is its dependence on interaction history. Performance consequently deteriorates for users or items with limited histories; the complete cold-start case occurs when no prior interaction is available.

Content-based filtering works by predicting whether a user would like an item based on the characteristics of the item itself. Items are described by their attributes, and the recommender system learns a score representing how well a particular item matches a user's preferences. Unlike collaborative filtering, content-based filtering can still recommend items even when interaction history is sparse, since it relies on item attributes rather than past interactions. Some classic techniques use keywords and TF-IDF of textual descriptions as the item attributes~\cite{kompan2010content, sohn2013contents}. With the rise of deep learning, item attributes are commonly represented as dense embedding vectors capturing multiple modalities~\cite{zhang2022latent}. In e-commerce platforms, the most naturally available and informative item attributes are textual descriptions, images, and user reviews. While textual descriptions and images characterize the product independently of any user experience, reviews are directly tied to user-item interactions: they reflect what users actually think about an item after purchasing or using it.

These two paradigms are not mutually exclusive. Modern recommendation methods increasingly combine both collaborative signals and semantic content to overcome their individual limitations~\cite{hu2025modality, li2024graph}. This hybrid direction motivates our work, which jointly leverages user-item interaction graphs and multimodal item representations.

\subsection{Graph-based Recommendation}
Graph neural networks (GNNs) have demonstrated strong performance in recommender systems by modeling user--item interactions as a bipartite graph and propagating collaborative signals through higher-order neighborhoods. By iteratively aggregating information from neighboring nodes, GNN-based methods capture complex interaction patterns that traditional collaborative-filtering approaches may fail to model. LightGCN~\cite{he2020lightgcn} reduces model complexity by removing nonlinear transformations and feature-transformation matrices, retaining only neighborhood aggregation. Despite its simplicity, it provides strong recommendation performance and has become a widely adopted baseline. However, such methods typically operate on homogeneous graphs with untyped edges and treat all interactions equally. They therefore fail to exploit the semantic information carried by ratings and reviews associated with user--item interactions. This limitation motivates heterogeneous graph structures with typed and semantically enriched edges.

\subsection{Review-based Recommendation}
User reviews represent a valuable source of information for recommendation, as they contain detailed opinions about specific item aspects and directly reflect user preferences. Early studies employed deep text encoders to extract semantic information from reviews, including DeepCoNN~\cite{deepconn}, which uses parallel TextCNNs to capture semantic features, and NARRE~\cite{narre}, which leverages attention to score the usefulness of each review. However, these methods treat reviews as static text features, failing to capture the fine-grained aspect-level preferences they contain. More recently, LLMs have been leveraged to extract structured (aspect, opinion, sentiment) triplets from reviews, enabling richer semantic modeling at the aspect level~\cite{HADSF}.

Beyond encoding reviews, a key question is how to integrate them into the recommendation system model. TGNN~\cite{shuai2023tgnn} proposes to incorporate reviews directly into the graph structure, using review embeddings as attention weights on the edges of a heterogeneous user--item graph. This allows review-dependent weights to influence the aggregated representations. However, TGNN is optimized for rating prediction rather than ranking. In real-world recommendation scenarios, the objective is to produce a ranked list of relevant items. Whether a product is assigned a score of $4.5$ or $2.5$ is less important than whether that relevant products are ranked above irrelevant ones. This misalignment motivates us to inherit the heterogeneous graph structure of TGNN while replacing the loss function with the Bayesian Personalized Ranking (BPR) loss which optimizes for the ranking instead of the score.

\subsection{Multimodal Recommendation}
Early multimodal recommendation methods demonstrated that incorporating product features can improve recommendation accuracy beyond collaborative signals alone. VBPR~\cite{vbpr} pioneered this direction by integrating visual features into the Bayesian Personalized Ranking framework, showing that visual item content provides meaningful signals for preference modeling. Building on this idea, BM3~\cite{zhou2023bm3} proposes a self-supervised framework that projects image and text features into a shared latent space and aligns them with collaborative signals. CFMM~\cite{cfmm} further introduces a softmax-based attention mechanism to weight image and text contributions dynamically for each item, together with modality-specific $k$-nearest-neighbor item graphs that capture inter-item similarities. Printf~\cite{lin2023printf} integrates all three modalities---item text, item images, and user reviews---but encodes reviews separately as user features through a dimension-wise attention mechanism.

\subsection{Collaborative-Semantic Alignment}
A key challenge in multimodal recommendation is bridging the gap between collaborative and semantic representation spaces. Recent work has begun to address this issue from different angles. Learning Item Representations Directly from Multimodal Features~\cite{lird} observes that item ID embeddings are initialized randomly while multimodal features are already informative from the start of training, causing the two spaces to diverge. They propose to initialize item embeddings directly from multimodal representations to reduce this gap from the outset. CFMM~\cite{cfmm} introduces a fusion loss inspired by Bayesian Personalized Ranking that encourages multimodal item representations to reflect user-preference rankings, ensuring that semantic features remain useful for recommendation beyond their role as auxiliary signals. This ranking objective is applied uniformly across items. GateSID~\cite{gatesid} instead uses a gate-regulated contrastive mechanism to adapt semantic--behavioral alignment to item popularity. Our approach combines a uniformly applied multimodal ranking loss with a separate popularity-weighted alignment objective that places a stronger constraint on sparse items.


\section{Methodology}

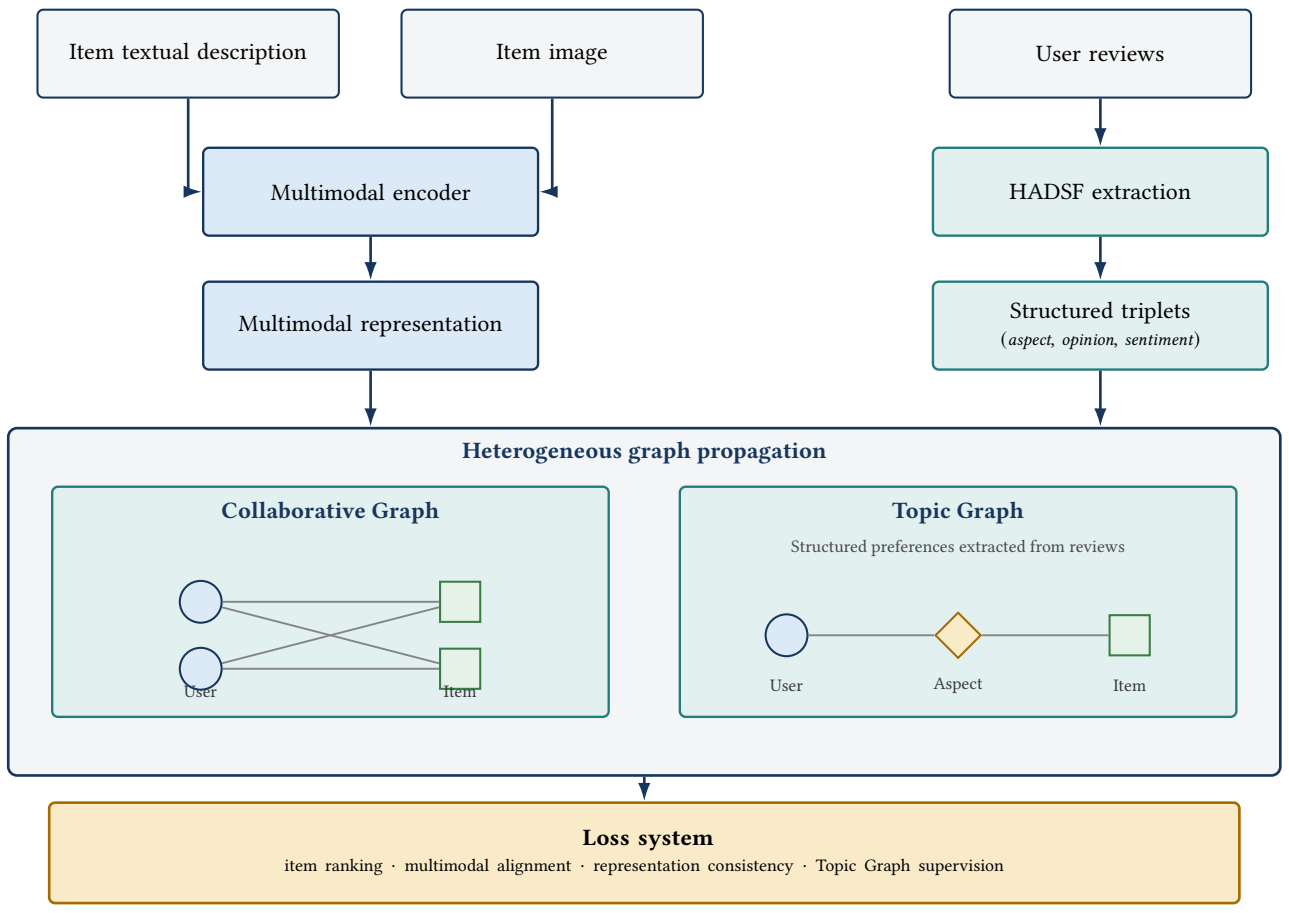
\begin{figure}[H]
\centering
\resizebox{\linewidth}{!}{%
\begin{tikzpicture}[
    font=\sffamily,
    input/.style={draw=diagramnavy, rounded corners=2pt,
                  fill=diagramgray, line width=0.7pt,
                  text width=3.25cm, minimum height=1.05cm,
                  inner sep=5pt, align=center, font=\small},
    process/.style={draw=diagramnavy, rounded corners=2pt,
                    fill=diagramblue, line width=0.8pt,
                    text width=3.65cm, minimum height=1.05cm,
                    inner sep=5pt, align=center, font=\small},
    reviewprocess/.style={process, draw=diagramteal,
                          fill=diagramteallight},
    group/.style={draw=diagramnavy, rounded corners=3pt,
                  fill=diagramgray, line width=0.9pt,
                  minimum width=15.2cm, minimum height=4.15cm},
    graphbox/.style={draw=diagramteal, rounded corners=2pt,
                     fill=diagramteallight, line width=0.8pt,
                     minimum width=6.65cm, minimum height=2.75cm},
    loss/.style={draw=diagramgold, rounded corners=2pt,
                 fill=diagramgoldlight, line width=0.9pt,
                 text width=13.8cm, minimum height=1.2cm,
                 inner sep=6pt, align=center, font=\small},
    title/.style={font=\small\bfseries, text=diagramnavy},
    subtitle/.style={font=\scriptsize, text=black!72, align=center},
    user/.style={circle, draw=diagramnavy, fill=diagramblue,
                 line width=0.7pt, minimum size=0.5cm, inner sep=0pt},
    item/.style={rectangle, draw=diagramgreen, fill=diagramgreenlight,
                 line width=0.7pt, minimum size=0.48cm, inner sep=0pt},
    aspect/.style={diamond, draw=diagramgold, fill=diagramgoldlight,
                   line width=0.7pt, minimum size=0.55cm, inner sep=0pt},
    arrow/.style={-{Latex[length=2.2mm,width=1.5mm]},
                  draw=diagramnavy, line width=0.9pt},
    edge/.style={draw=black!48, line width=0.65pt},
    legend/.style={font=\scriptsize, text=black!78}
]

\node[input] (texte) at (0,0) {Item textual description};
\node[input] (image) at (4.35,0) {Item image};
\node[input] (avis) at (10.9,0) {User reviews};

\node[process] (encodeur) at (2.18,-1.65)
  {Multimodal encoder\\[-1pt]\scriptsize};
\node[process] (hmodal) at (2.18,-3.25)
  {Multimodal representation};

\node[reviewprocess] (hadsf) at (10.9,-1.65)
  {HADSF extraction};
\node[reviewprocess] (triplet) at (10.9,-3.25)
  {Structured triplets\\[-1pt]
   \scriptsize$(\textit{aspect},\,\textit{opinion},\,\textit{sentiment})$};

\node[group] (outer) at (5.45,-6.55) {};
\node[title, fill=diagramgray, inner xsep=6pt]
  at ([yshift=-0.3cm]outer.north) {Heterogeneous graph propagation};

\node[graphbox] (graphcollab) at (1.7,-6.55) {};
\node[title, fill=diagramteallight, inner xsep=4pt]
  at ([yshift=-0.32cm]graphcollab.north) {Collaborative Graph};

\node[graphbox] (graphaspect) at (9.2,-6.55) {};
\node[title, fill=diagramteallight, inner xsep=4pt]
  at ([yshift=-0.32cm]graphaspect.north) {Topic Graph};
\node[subtitle] at ([yshift=-0.75cm]graphaspect.north)
  {Structured preferences extracted from reviews};

\node[user] (u1) at (0.15,-6.55) {};
\node[user] (u2) at (0.15,-7.35) {};
\node[item] (i1) at (3.25,-6.55) {};
\node[item] (i2) at (3.25,-7.35) {};
\draw[edge] (u1) -- (i1);
\draw[edge] (u1) -- (i2);
\draw[edge] (u2) -- (i1);
\draw[edge] (u2) -- (i2);
\node[legend] at (0.15,-7.62) {User};
\node[legend] at (3.25,-7.62) {Item};

\node[user] (ua) at (7.15,-6.95) {};
\node[aspect] (ta) at (9.2,-6.95) {};
\node[item] (ia) at (11.25,-6.95) {};
\draw[edge] (ua) -- (ta);
\draw[edge] (ta) -- (ia);
\node[legend] at (7.15,-7.55) {User};
\node[legend] at (9.2,-7.55) {Aspect};
\node[legend] at (11.25,-7.55) {Item};

\node[loss] (loss) at (5.45,-9.55)
  {\textbf{ Loss system}\\[-1pt]
   \scriptsize item ranking $\cdot$ multimodal alignment $\cdot$ representation consistency
   $\cdot$ Topic Graph supervision};

\draw[arrow] (texte.south) |- (encodeur.west);
\draw[arrow] (image.south) |- (encodeur.east);
\draw[arrow] (avis) -- (hadsf);
\draw[arrow] (encodeur) -- (hmodal);
\draw[arrow] (hadsf) -- (triplet);
\draw[arrow] (hmodal.south) -- ([xshift=-3.27cm]outer.north);
\draw[arrow] (triplet.south) -- ([xshift=5.45cm]outer.north);
\draw[arrow] (outer) -- (loss);

\end{tikzpicture}
}
\caption{Overview of the proposed architecture.}
\Description{The architecture processes item text and images with a Modal Encoder and reviews with HADSF. A Collaborative Graph and a Topic Graph then produce representations optimized by a joint loss system for top-k recommendation.}
\label{fig:architecture}
\end{figure}

Figure~\ref{fig:architecture} illustrates the overall architecture of the proposed framework. Given users, items, and their associated multimodal content, the model processes three complementary modalities and integrates them through two parallel graph encoders and an adaptive loss system to produce top-$K$ recommendations.

For the review modality, raw user reviews are processed by an LLM-based extraction mechanism to derive structured (aspect, opinion, sentiment) triplets, capturing fine-grained user preferences beyond simple rating scores. The extracted triplets and their embeddings serve as inputs to the graph encoders. For the visual and textual item modalities, item images and textual descriptions are independently encoded by pretrained encoders and fed into a Modal Encoder, which fuses them via learned per-item attention weights. An image--text alignment loss encourages the two modalities of the same item to be mapped to nearby representations in the shared latent space.

The model employs two parallel graph encoders operating on heterogeneous graph structures. The Collaborative Graph represents each interaction by bidirectional user--item edges. Review information is used to weight message passing and is injected into the second propagation layer, producing collaborative user and item representations. In parallel, the Topic Graph connects review triplets, aspect nodes, users, and items to model fine-grained preferences at the aspect level.

The model is trained with a BPR ranking loss, an image--text alignment loss, a popularity-weighted collaborative--multimodal alignment loss, a ranking loss applied to the multimodal item representations, and a mutual-information objective that retrieves the triplet associated with each user--item interaction. The resulting relevance scores are used to rank candidate items and produce top-$K$ recommendations.

\subsection{Review Processing}

Following HADSF~\cite{HADSF}, we adopt a two-stage LLM-based pipeline, implemented with Llama~3.1~\cite{dubey2024llama}, to extract structured semantic information from raw user reviews.

In the first stage, reviews are recursively summarized at the item level. The LLM extracts candidate themes from these summaries, after which semantically similar themes are clustered to construct a global aspect vocabulary $\mathcal{A}^{*}$. This vocabulary provides a consistent set of aspects across users and items.

In the second stage, each review is processed together with $\mathcal{A}^{*}$ and contextual information obtained from the corresponding user's review history and the item's review history. The LLM extracts one or more $(\text{aspect},\text{opinion},\text{sentiment})$ triplets. The complete review and each extracted triplet are encoded separately with a pretrained BERT encoder and projected to 128 dimensions. The complete-review representation $\mathbf{r}_{ui}$ is used as a user--item edge feature in the Collaborative Graph, whereas the triplet representation $\mathbf{t}_{ui}$ is used as a node feature and supervision target in the Topic Graph.

\subsection{Modal Encoder}
To obtain a unified item representation from visual and textual modalities, we propose a Modal Encoder that fuses pre-extracted image and text features into a shared semantic space. Item images are encoded by a pretrained ResNet50, and textual item descriptions are encoded by a pretrained MiniLM SentenceTransformer. Both encoders are frozen during training — only the projection layers are learned.

Each modality is projected into a shared latent space via a linear transformation followed by a nonlinear activation and layer normalization, producing modality-specific representations $\mathbf{h}_v$ and $\mathbf{h}_t$ for image and text, respectively.

Inspired by CFMM~\cite{cfmm}, we adopt a per-item attention mechanism to adaptively fuse the two modalities. For each item, attention weights are computed over the image and text representations and normalized via softmax:
\begin{equation}
[w_{img}, w_{txt}] = \text{Softmax}([\text{Query}(\mathbf{h}_v),
\text{Query}(\mathbf{h}_t)])
\end{equation}
The final fused representation is then obtained as:
\begin{equation}
\mathbf{h}_{modal} = w_{img} \cdot \mathbf{h}_v + w_{txt} \cdot \mathbf{h}_t
\end{equation}
This per-item weighting allows the model to learn different image and text contributions for different items. The learned weights indicate which projected modality is most useful to the training objective; they do not, by themselves, measure the intrinsic quality or completeness of either modality.

To enforce semantic coherence between the two modalities, we apply an InfoNCE contrastive loss between $\mathbf{h}_v$ and $\mathbf{h}_t$, detailed in Section~\ref{sec:learning_and_optimization}. This objective encourages the projected image and text representations of the same item to be close in the shared latent space before they are fused into $\mathbf{h}_{modal}$. The fused modal representation is also used to initialize the item embedding before graph training, reducing the initial discrepancy between semantic and collaborative representations.

\subsection{Graph Encoding}
The proposed framework relies on two parallel heterogeneous graphs, both adapted from TGNN~\cite{shuai2023tgnn}. The first, referred to as the \textbf{Collaborative Graph}, models user-item interactions enriched with review embeddings as edge features. The second, referred to as the \textbf{Topic Graph}, captures fine-grained semantic preferences at the aspect level through topic nodes derived from the LLM extraction pipeline. We adapt both graphs to a ranking framework by replacing the original rating prediction objective with a BPR ranking loss. Both graphs share the same user and item nodes but differ in their edge structure and encoding mechanism.

\subsubsection{Collaborative Graph Encoding}
The collaborative graph connects users and items through bidirectional interaction edges, where each edge carries the BERT embedding of the associated review as a feature. For each interaction edge $(u,i)$, a learnable scalar $p_{ui}$ is initialized from the review embedding and updated during training. The first-layer message from user $u$ to item $i$ is:
\begin{equation}
\mathbf{m}_{u \rightarrow i}^{(1)} = p_{ui}\mathbf{h}_u^{(0)}
\end{equation}
The learned coefficient controls the contribution of the interaction to message passing; it should not be interpreted as an independently validated measure of review informativeness.

In the second GCN layer, the review embedding is additionally injected directly into the message:
\begin{equation}
\mathbf{m}_{u \rightarrow i}^{(2)} = p_{ui}\mathbf{h}_u^{(1)}
+ s_{ui}\mathbf{r}_{ui}
\end{equation}
where $\mathbf{r}_{ui}$ is the review representation and $s_{ui}$ is a learned coefficient controlling its contribution. After two GCN layers with symmetric normalization, the final collaborative representations $\mathbf{u}_{rf} \in \mathbb{R}^{128}$ and $\mathbf{i}_{rf} \in \mathbb{R}^{128}$ are obtained for users and items, respectively.

\subsubsection{Topic Graph Encoding}
The topic graph connects review-triplet nodes to user- and item-specific aspect nodes, which are in turn connected to user and item nodes. This structure allows the model to capture fine-grained semantic preferences at the aspect level, going beyond the binary signal of whether an interaction occurred. The topic graph produces topic-level representations $\mathbf{u}_{tf}$ and $\mathbf{i}_{tf}$ for users and items, respectively, which are combined with the collaborative representations in the optimization objective.

\subsection{Learning and Optimization} \label{sec:learning_and_optimization}
\subsubsection{BPR Loss}
The primary training objective is Bayesian Personalized Ranking (BPR)~\cite{rendle2009bpr}, which directly optimizes the relative ranking of items. For each user $u$, the model learns to score a positively interacted item $i^+$ higher than an unobserved item $i^-$:
\begin{equation}
\mathcal{L}_{BPR} = -\frac{1}{|\mathcal{B}|} \sum_{(u,i^+,i^-)
\in \mathcal{B}} \log \sigma(\hat{y}_{u,i^+} - \hat{y}_{u,i^-})
\end{equation}
where $\hat{y}_{u,i}=\mathbf{u}_{r}\cdot\mathbf{i}_{r}$ is the score computed from the Collaborative Graph representations and $\sigma$ is the sigmoid function. An L2 regularization term is added to prevent overfitting.

\subsubsection{Modal Alignment Loss}
To align the visual and textual representations of the same item, we apply a batch-wise cross-entropy contrastive loss between $\mathbf{h}_v$ and $\mathbf{h}_t$. For each item in the batch, its image representation must be closer to its own text representation than to those of all other items in the batch:
\begin{equation}
\mathcal{L}_{modal} = -\frac{1}{|\mathcal{B}|} \sum_{i \in
\mathcal{B}} \log \frac{\exp(\mathbf{h}_{v,i} \cdot \mathbf{h}_{t,i})}
{\sum_{j \in \mathcal{B}} \exp(\mathbf{h}_{v,i} \cdot
\mathbf{h}_{t,j})}
\end{equation}
where $\mathbf{h}_{v,i}$ and $\mathbf{h}_{t,i}$ are the image and text representations of item $i$. This objective encourages the image and text representations of the same item to be close in the shared latent space.

\subsubsection{Modality Gap Loss}
To bridge the gap between collaborative and semantic representation spaces, we introduce a contrastive alignment loss that pulls item embeddings closer to their modal representations $\mathbf{h}_{modal}$:
\begin{equation}
\mathcal{L}_{MG} = \frac{1}{|\mathcal{B}|}
\sum_{i \in \mathcal{B}} -(1 - \alpha_i)
\log \frac{\exp(\mathbf{h}^{(0)}_i \cdot \mathbf{h}_{modal,i})}
{\sum_{j \in \mathcal{B}} \exp(\mathbf{h}^{(0)}_i \cdot
\mathbf{h}_{modal,j})}
\end{equation}
This loss is weighted by $(1 - \alpha_i)$, where $\alpha_i$ is the log-normalized training degree:
\begin{equation}
\alpha_i = \frac{\log(1+\deg(i))}{\log(1+\deg_{\max})}.
\end{equation}
Items with few interactions contribute more to this alignment, ensuring that their collaborative representations remain anchored to the semantically rich modal signal.

\subsubsection{Fusion Loss}
Inspired by CFMM~\cite{cfmm}, this loss ensures that the modal representations $\mathbf{h}_{modal}$ remain directly predictive of user preferences rather than serving only as an alignment target. We apply a BPR objective directly on $\mathbf{h}_{modal}$:
\begin{equation}
\mathcal{L}_{F} = \frac{1}{|\mathcal{B}|}
\sum_{(u,i^+,i^-) \in \mathcal{B}}
-\log \sigma(\mathbf{u}_{r} \cdot \mathbf{h}_{modal,i^+} -
\mathbf{u}_{r} \cdot \mathbf{h}_{modal,i^-})
\end{equation}
where the objective is applied uniformly to the sampled training triples. Unlike the Modality Gap Loss, the Fusion Loss is not weighted by item popularity; it makes the modal representations directly predictive of the ranking objective for all items.

\subsubsection{MI Loss}
To supervise the Topic Graph, collaborative and topic representations are first combined as $\widetilde{\mathbf{u}}=\mathbf{u}_{r}+\mathbf{u}_{t}$ and $\widetilde{\mathbf{i}}=\mathbf{i}_{r}+\mathbf{i}_{t}$. Their concatenation is projected into an interaction representation $\mathbf{th}_{u,i}=f([\widetilde{\mathbf{u}}\Vert \widetilde{\mathbf{i}}])$. The mutual-information objective then retrieves the review triplet associated with each user--item interaction from the triplets of the other interactions in the batch:
\begin{equation}
\mathcal{L}_{MI} = -\frac{1}{|\mathcal{B}|} \sum_{(u,i)
\in \mathcal{B}} \log \frac{\exp(\mathbf{th}_{u,i} \cdot
\mathbf{t}_{u,i})}{\sum_{(u',i') \in \mathcal{B}}
\exp(\mathbf{th}_{u,i} \cdot \mathbf{t}_{u',i'})}
\end{equation}
where $\mathbf{t}_{u,i}$ is the embedding of the corresponding extracted triplet. To avoid interfering with the primary ranking objective, this loss is introduced gradually via a linear warmup over the first training epochs.

The complete training objective is:
\begin{equation}
\mathcal{L} = \mathcal{L}_{BPR}
+ \lambda_{modal}\mathcal{L}_{modal}
+ \lambda_{mg}\mathcal{L}_{MG}
+ \mathcal{L}_{F}
+ \lambda_{MI}\mathcal{L}_{MI}.
\end{equation}
Only $\mathcal{L}_{MG}$ contains the item-dependent factor $(1-\alpha_i)$.


\section{Experimental Settings}
\subsection{Datasets}

We evaluate our model on two product categories from the Amazon Reviews 2023 dataset~\cite{hou2024amazon}: Musical Instruments and Baby. Both categories provide product images, textual descriptions, and user reviews simultaneously. Since our framework requires each user-item interaction to be associated with a review, we retain only interactions that include a review, ensuring that each edge in the heterogeneous graph carries a review embedding.

Each dataset is split into 80\% for training, 10\% for validation, and 10\% for testing. To ensure a meaningful evaluation, we retain only users and items with at least one interaction in the training set, guaranteeing that all retained nodes have a collaborative signal during training. The dataset statistics are summarized in Table~\ref{tab:datasets}.

\begin{table}[h]
\caption{Dataset statistics.}
\label{tab:datasets}
\resizebox{\columnwidth}{!}{
\begin{tabular}{lcc}
\toprule
& \textbf{Musical Instruments} & \textbf{Baby} \\
\midrule
Users         & 17 318 & 25 474  \\
Items         & 4 886  & 5 492  \\
Interactions  & 58 469   & 70 205 \\
Sparsity      & 99.93\% & 99.95\% \\\bottomrule
\end{tabular}
}
\end{table}

\subsection{Evaluation Protocol and Metrics}
\subsubsection{Item Ranking}
We use Recall@$K$ and normalized Discounted Cumulative Gain (nDCG@$K$), with $K=10$. Recall@$K$ measures the fraction of relevant items retrieved in the top-$K$ list, while nDCG@K additionally accounts for the ranking position of relevant items, assigning higher scores to relevant items appearing earlier in the list.

For each test interaction, the observed item is ranked against 99 items sampled from those absent from the user's interaction history. Because each candidate set contains one positive item, Recall@10 indicates whether that item appears in the top 10, whereas nDCG@10 additionally rewards a higher rank within the top 10.

This sampled protocol reduces evaluation cost and uses the same negative candidates for all models, but it does not constitute an exhaustive ranking over the complete item catalog.

\subsubsection{HADSF Extraction Quality}
We evaluate the reliability of the extracted triplets using two metrics. The Aspect Drift Rate (ADR) is the proportion of extracted aspects that do not belong to the global aspect vocabulary $\mathcal{A}^{*}$; lower values indicate less aspect drift. The Opinion Fidelity Rate (OFR) is the average cosine similarity between an extracted opinion and its source review; higher values indicate greater semantic consistency with the original review.

\subsubsection{Triplet Ranking}
To evaluate the Topic Graph independently, the topic representations $\mathbf{u}_{t}$ and $\mathbf{i}_{t}$ are concatenated and projected to obtain $\mathbf{s}^{t}_{ui}=f_t([\mathbf{u}_{t}\Vert \mathbf{i}_{t}])$. This representation is compared with the triplet embedding of every test interaction. The target triplet is treated as positive and all other test triplets as negatives; Recall@10 and nDCG@10 are then computed from the resulting ranking.

Items are further divided into two groups based on their interaction density: items with fewer than 10 interactions are considered \textit{Sparse items}, while items with 10 or more interactions are considered \textit{popular items}. These sparse items have limited but nonzero training histories and therefore differ from the complete cold-start setting. Performance is reported separately for each group, allowing a fine-grained analysis of each model's ability to handle items of varying popularity.

\subsection{Implementation Details}

All experiments are implemented in PyTorch and run on a GPU cluster. For item visual features, we use pre-extracted ResNet50 average pooling features projected to 128 dimensions via a single-layer MLP. For textual item features, we use pre-extracted embeddings from all-MiniLM-L6-v2 (384-dim), similarly projected to 128 dimensions. Both feature extractors are frozen during training — only the projection layers are learned.

The heterogeneous user--item graph is constructed with two directed edge types: (user $\rightarrow$ item) and (item $\rightarrow$ user), both enriched with the corresponding review embedding. The Collaborative Graph uses two layers and an embedding dimension of 128.

For optimization, we use Adam with a learning rate of $0.001$, L2 regularization of $10^{-4}$, and dropout of $0.8$. The batch size is set to 512 with $n_{neg}=5$ negative items sampled per positive interaction. Loss weights are set to $\lambda_{modal}=0.1$, $\lambda_{mg}=0.1$, and $\lambda_{MI}=0.01$ with a linear warmup of 100 epochs. Training uses early stopping with a patience of 100 epochs.

Unless explicitly stated otherwise, MPIUR refers to the configuration using ResNet50 as the visual encoder. This configuration is used for the global, group-wise, ablation, and hyperparameter results; the MLLM variants are evaluated separately.

\section{Results}
\subsection{Overall Performance}
\begin{table*}[h]
\caption{Performance comparison in terms of nDCG@10 and Recall@10 across all item groups.}
\label{tab:overall}
\resizebox{\textwidth}{!}{
\begin{tabular}{lcccccccccccc}
\toprule
& \multicolumn{6}{c}{\textbf{Musical Instruments}}
& \multicolumn{6}{c}{\textbf{Baby}} \\
\cmidrule(lr){2-7} \cmidrule(lr){8-13}
& \multicolumn{2}{c}{\textbf{Global}}
& \multicolumn{2}{c}{\textbf{Sparse items}}
& \multicolumn{2}{c}{\textbf{Popular items}}
& \multicolumn{2}{c}{\textbf{Global}}
& \multicolumn{2}{c}{\textbf{Sparse items}}
& \multicolumn{2}{c}{\textbf{Popular items}} \\
\cmidrule(lr){2-3} \cmidrule(lr){4-5} \cmidrule(lr){6-7}
\cmidrule(lr){8-9} \cmidrule(lr){10-11} \cmidrule(lr){12-13}
\textbf{Model}
& N@10 & R@10
& N@10 & R@10
& N@10 & R@10
& N@10 & R@10
& N@10 & R@10
& N@10 & R@10 \\
\midrule

LightGCN
& 0.1613 & 0.2577
& 0.1142 & 0.1943
& 0.2217 & 0.3389
& 0.1636 & 0.2984
& 0.1230 & 0.2255
& 0.2013 & 0.3660 \\

TGNN
& 0.2249 & 0.3990
& 0.2289 & 0.4152
& 0.2198 & 0.3782
& 0.1977 & 0.3538
& 0.1941 & 0.3498
& 0.2010 & 0.3575 \\

BM3
& 0.2276 & 0.3955
& 0.2183 & 0.3870
& \textbf{0.2395} & \textbf{0.4065}
& 0.2025 & 0.3859
& 0.2048 & 0.3976
& 0.2004 & \textbf{0.3751} \\

MPIUR
& \textbf{0.2344} & \textbf{0.4042}
& \textbf{0.2347} & \textbf{0.4153}
& 0.2340 & 0.3900
& \textbf{0.2194} & \textbf{0.3891}
& \textbf{0.2384} & \textbf{0.4161}
& \textbf{0.2018} & 0.3640 \\

\bottomrule
\end{tabular}
}
\end{table*}

Table~\ref{tab:overall} compares MPIUR with the three baselines globally and within the sparse- and popular-item groups.

\textbf{Comparison with LightGCN.} LightGCN relies only on the interaction structure. MPIUR obtains substantially higher global scores on both datasets. The largest nDCG@10 differences occur for sparse items: $+0.1205$ on Musical Instruments and $+0.1154$ on Baby. MPIUR also performs better on popular items except for Recall@10 on Baby, where LightGCN obtains 0.3660 versus 0.3640.

\textbf{Comparison with TGNN.} TGNN is the direct graph backbone: it already uses reviews as edge features but does not include the visual and textual item modalities or the proposed loss system. We replace the original MSE loss function with the BPR loss function. MPIUR obtains higher global scores on both datasets. On sparse Musical Instruments items, Recall@10 is essentially unchanged (0.4153 versus 0.4152), while nDCG@10 increases from 0.2289 to 0.2347. The improvements are larger on sparse Baby items, where MPIUR reaches 0.2384 nDCG@10 and 0.4161 Recall@10, compared with 0.1941 and 0.3498 for TGNN. MPIUR also remains slightly better on the popular-item metrics.

\textbf{Comparison with BM3.} BM3 uses visual and textual item features but does not incorporate review content. MPIUR achieves higher global nDCG@10 and Recall@10 on both datasets and performs better on all sparse-item metrics. The pattern changes for popular items: BM3 performs better on both Musical Instruments metrics and on Baby Recall@10, whereas MPIUR obtains the higher Baby nDCG@10.

\subsection{Review Extraction and Topic Graph Evaluation} \label{sec:review_ext_tg_eval}
Table~\ref{tab:hadsf-quality} reports the quality of the triplets extracted by HADSF. The ADR is 0.0125 on Musical Instruments and 0.0126 on Baby. The corresponding OFR values are 0.9358 and 0.9375.

\begin{table}[h]
\centering
\caption{Quality of the HADSF extraction.}
\label{tab:hadsf-quality}
\begin{tabular}{lcc}
\toprule
\textbf{Dataset} & \textbf{OFR} & \textbf{ADR} \\
\midrule
Musical Instruments & 0.9358 & 0.0125 \\
Baby                 & 0.9375 & 0.0126 \\
\bottomrule
\end{tabular}
\end{table}

Table~\ref{tab:triplet-ranking} evaluates the Topic Graph on triplet ranking against approximately 5,200 candidates per interaction. For a random ranking over a candidate pool of this size, the expected Recall@10 is approximately 0.0019 and the expected nDCG@10 is approximately 0.00087. Musical Instruments obtains 0.0527 Recall@10 and 0.0314 nDCG@10, while Baby obtains 0.0601 and 0.0343, respectively. Both datasets therefore score substantially above the random-ranking baseline and obtain similar values. An error analysis shows that, in approximately 65\% of failed cases, the model retrieves the correct product aspect but assigns a different opinion to it; the remaining errors mainly confuse different aspects of the same product.

\begin{table}[h]
\centering
\caption{Topic Graph performance on triplet ranking.}
\label{tab:triplet-ranking}
\begin{tabular}{lcc}
\toprule
\textbf{Dataset} & \textbf{Recall@10} & \textbf{nDCG@10} \\
\midrule
Musical Instruments & 0.0527 & 0.0314 \\
Baby                 & 0.0601 & 0.0343 \\
\midrule
Random (expected) & 0.0019 & 00087 \\
\bottomrule
\end{tabular}
\end{table}

\subsection{Ablation Studies}

\begin{table}[!ht]
\centering
\caption{Ablation study of the proposed model in terms of nDCG@10 on the ``Musical Instruments'' dataset.}
\label{tab:ablation}
\resizebox{\columnwidth}{!}{
\begin{tabular}{lccc}
\toprule
\textbf{Variant} & \textbf{Global} & \textbf{Sparse Items} & \textbf{Popular Items} \\
\midrule
Full model              & \textbf{0.2344} & \textbf{0.2347} & \textbf{0.2340} \\
w/o Modality Gap Loss   & 0.2135 & 0.2126 & 0.2147 \\
w/o Fusion Loss         & 0.2150 & 0.2130 & 0.2176 \\
w/o Image \& Text Features & 0.2177 & 0.2176 & 0.2178 \\
w/o Review Features     & 0.1972 & 0.1869 & 0.2104 \\
w/o All Modalities      & 0.1919 & 0.1814 & 0.2053 \\
\bottomrule
\end{tabular}
}
\end{table}

To assess the contribution of each component, we conduct an ablation study by removing one component at a time from the full model on the ``Musical Instruments'' dataset. The variants are defined as follows:

\begin{itemize}

\item \textbf{w/o Modality Gap Loss}: the contrastive alignment between item embeddings and modal representations is removed.

\item \textbf{w/o Fusion Loss}: the BPR objective applied directly on modal representations is removed, meaning $\mathbf{h}_{modal}$ is no longer explicitly guided toward the ranking objective.

\item \textbf{w/o Image \& Text Features}: image and text features are excluded, removing the Modal Encoder entirely, as well as the Fusion Loss, Modality Gap Loss, and Modal Alignment Loss.

\item \textbf{w/o Review Features}: review embeddings are removed from the graph edges; the Topic Graph and its associated MI Loss are also removed.

\item \textbf{w/o All Modalities}: all three modalities — image, text, and reviews — are removed simultaneously, reducing the model to a pure collaborative filtering baseline.

\end{itemize}

Table~\ref{tab:ablation} shows that removing review features produces the largest decrease among the individual modality removals in all three groups. The sparse-item score decreases from 0.2347 to 0.1869, whereas removing image and text features decreases it to 0.2176. Removing all modalities produces the lowest scores: 0.1919 globally, 0.1814 for sparse items, and 0.2053 for popular items.

Removing the Modality Gap Loss decreases nDCG@10 to 0.2135 globally, 0.2126 for sparse items, and 0.2147 for popular items. Removing the Fusion Loss yields 0.2150, 0.2130, and 0.2176, respectively. Thus, every tested ablation performs below the full model in every reported group.

\subsection{Modality Gap Analysis}
\begin{figure}[H]
    \centering
    \includegraphics[width=0.9\linewidth]{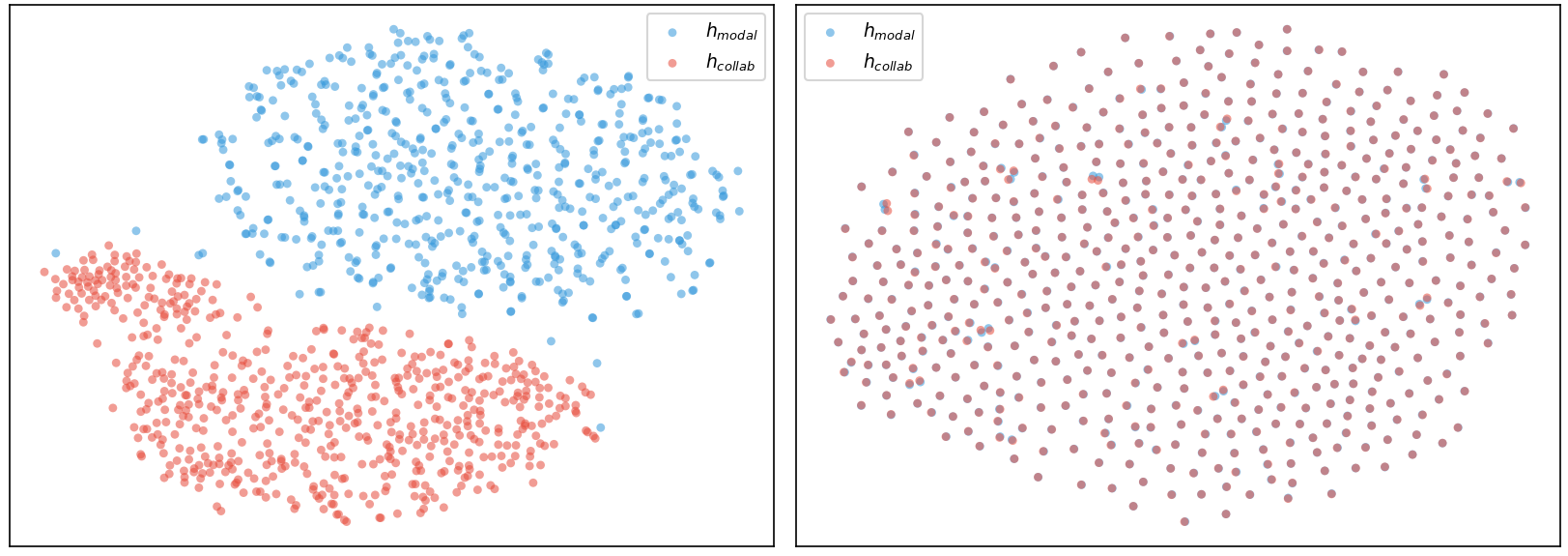}
    \caption{t-SNE visualization of collaborative and modal embeddings before and after alignment on a sample of the Musical Instruments dataset.}
    \Description{Two scatter plots compare collaborative and modal item embeddings. The two groups are separated before alignment and overlap substantially after alignment.}
    \label{fig:tsne}
\end{figure}

Figure~\ref{fig:tsne} illustrates the effect of our alignment strategy on the spatial distribution of collaborative and modal embeddings. Before alignment, the two types of representations occupy visibly distinct regions. After training, the two sets of points overlap more in the two-dimensional projection. This figure is a qualitative visualization and does not measure separation in the original high-dimensional space.

\subsection{Hyperparameter Sensitivity}

\subsubsection{Number of Negative Samples}
The number of negative samples per positive interaction $n_{neg}$ is tuned over a range from 1 to 64. We select $n_{neg}=5$ based on validation performance. The validation score reaches its maximum at five negatives and decreases for every tested value above five.
\begin{figure}[H]
    \centering
    \includegraphics[width=0.7\linewidth]{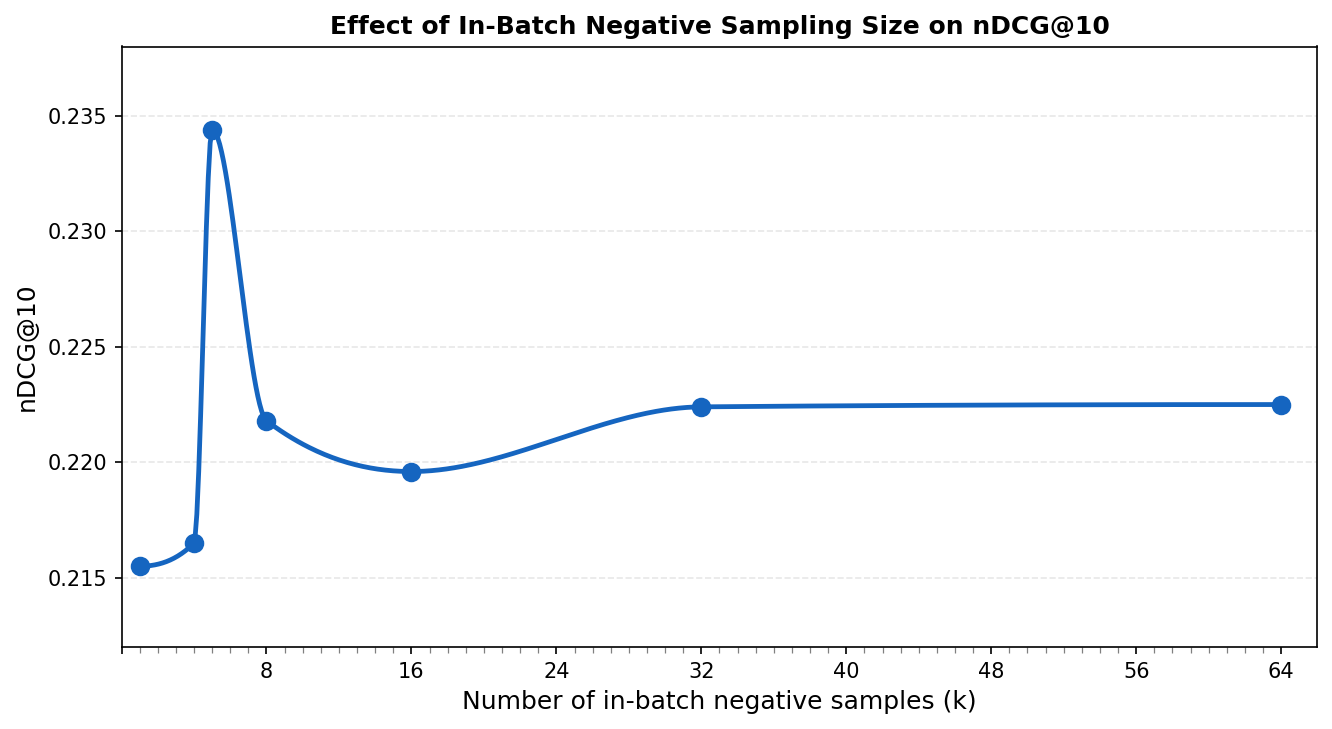}
    \caption{Sensitivity to the number of negative samples $k$ on the Musical Instruments validation set.}
    \Description{A line chart of validation nDCG at 10 as the number of negative samples varies from 1 to 64. The best value occurs at five negative samples.}
    \label{fig:negative-samples}
\end{figure}

\subsubsection{Sparse-Item Threshold Analysis}
\begin{figure}[H]
    \centering
    \includegraphics[width=0.7\linewidth]{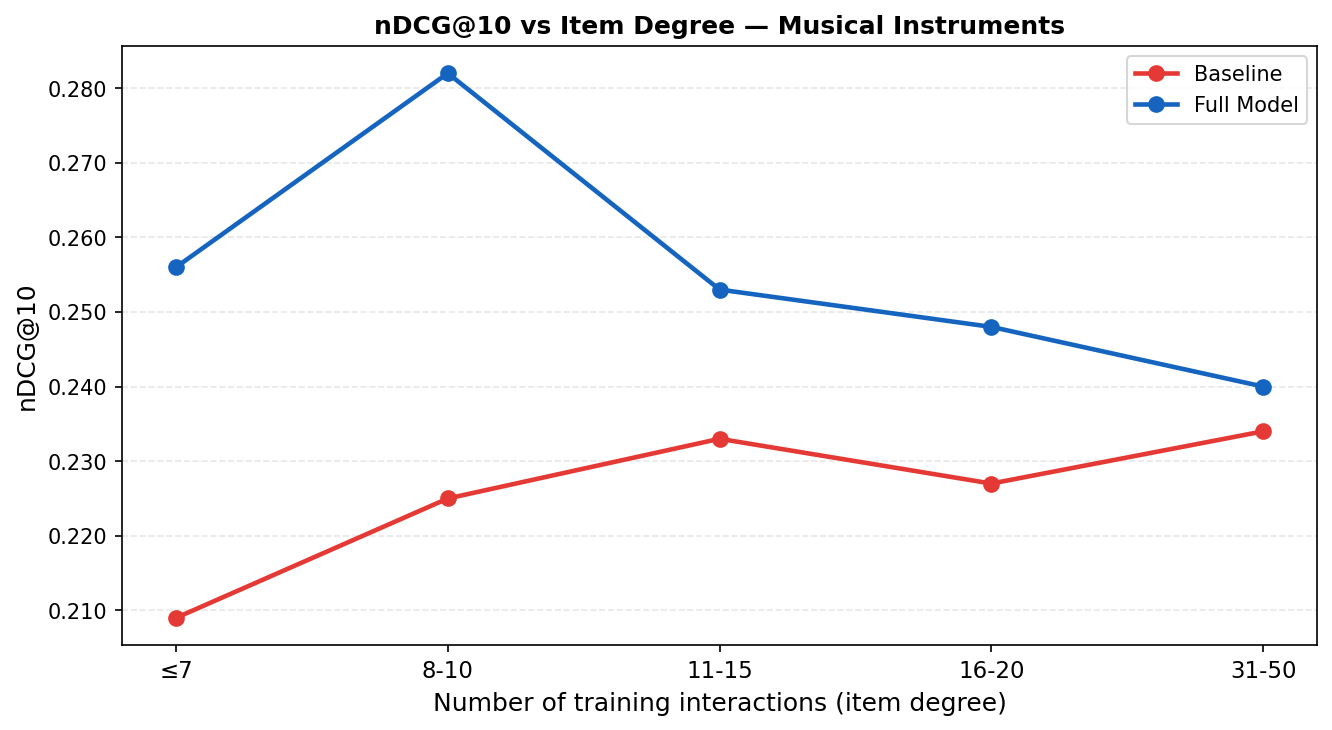}
    \caption{nDCG@10 of the full model and the variant without modalities across item-degree bins on Musical Instruments.}
    \Description{A line chart compares the full model with the variant without modalities across item interaction-degree bins. The full model performs better in every bin, with the largest gap in the 8--10 interaction bin.}
    \label{fig:threshold}
\end{figure}

Item interaction counts in our datasets range from 1 to nearly 500. We evaluate $T=10$ as the operational threshold for defining sparse items. Figure~\ref{fig:threshold} shows the nDCG@10 of our full model and its variant without any modality (w/o All Modalities) across item degree bins.

The full model outperforms the variant without modalities in every displayed bin. The largest observed gap occurs in the 8--10 bin. The gap is smaller in every bin above 10 interactions and reaches its minimum in the 31--50 bin. We use 10 interactions as the operational threshold for the group-wise evaluation.

\section{VLM-based Visual Encoding Experiment}
\label{sec:vlm-encoding}

Large Language Models (LLMs) and Multimodal LLMs (MLLMs) have demonstrated a strong ability to capture rich semantic information that pretrained encoders such as CNNs often fail to expose, particularly attributes that are visually implicit but not explicitly stated in item metadata~\cite{lird, ye2025mllmmsr}. Motivated by this, recent work has explored replacing traditional visual/textual feature extractors with MLLM-based pipelines to obtain item representations more directly aligned with the recommendation objective, rather than relying on generic embeddings~\cite{lird, ye2025mllmmsr}. Building on this direction, we investigate whether replacing the frozen CNN-based image encoder in our Modal Encoder with an MLLM can yield item representations that better capture recommendation-relevant visual semantics.

We replace the frozen ResNet50 image encoder with VLM-based pipelines using Qwen2.5-VL-7B-Instruct~\cite{qwen2025vl} and LLaVA-13B~\cite{liu2023llava}. Item images are first converted into textual descriptions, and the resulting descriptions are encoded with the same MiniLM text encoder used elsewhere in the framework. The remainder of the Modal Encoder architecture is unchanged.

\textbf{Prompt Design.} The prompt is designed to elicit visual attributes that are implicit in the product image but absent from the item's original textual metadata, allowing the generated description to serve as a complementary signal rather than a redundant one. To reduce hallucinations, the model is explicitly instructed to describe only information that can be inferred from the image itself. The prompt is further contextualized with the item's category (\{DATASET\}), guiding the model toward domain-specific attributes instead of producing a generic description. Finally, the prompt follows an instructional format by explicitly specifying the aspects to cover---core function, key features, target user, style and design, and quality and materials---to encourage consistent descriptions across items. The output is also limited to 100 words, producing concise captions suitable for downstream text encoding.
\begin{quote}
\small
\textit{Provide a brief product description based on the image, focusing on the key features and benefits relevant to the \{DATASET\} category.}

\begin{itemize}
    \item \textbf{Core Function:} What is the primary purpose of this product?
    \item \textbf{Key Features:} What are the standout features that set it apart from similar products?
    \item \textbf{Target User:} Who is the ideal user or buyer for this product?
    \item \textbf{Style and Design:} What are the product's overall appearance, color, and design?
    \item \textbf{Quality \& Materials:} What can be inferred about the materials or build quality?
\end{itemize}

\textit{Answer in one paragraph of no more than 100 words.}
\end{quote}

\begin{table}[h]
\caption{Global nDCG@10 comparison between CNN-based and VLM-based visual encoding.}
\label{tab:vlm}
\resizebox{\columnwidth}{!}{%
\begin{tabular}{lccc}
\toprule
\textbf{Dataset} & \textbf{ResNet50} & \textbf{Qwen2.5-VL} & \textbf{LLaVA-13B} \\
\midrule
Musical Instruments & 0.2344 & \textbf{0.2381} & 0.2316 \\
Baby                 & \textbf{0.2194} & 0.2113 & 0.2091 \\
\bottomrule
\end{tabular}
}
\end{table}

Table~\ref{tab:vlm} reports the results. On Musical Instruments, Qwen2.5-VL obtains a small improvement over ResNet50, whereas LLaVA-13B remains below the ResNet50 baseline. On Baby, ResNet50 outperforms both VLM variants. Qwen2.5-VL is better than LLaVA-13B on both datasets.

\section{Limitations}
Several limitations should be considered when interpreting the results. First, preprocessing retains only interactions associated with a textual review. Purchases, ratings, and other interactions without comments are excluded even though they contain useful collaborative information, which may bias the data toward users who write reviews frequently. Second, every retained item must have at least one training interaction; the model therefore addresses sparse items rather than the complete item cold-start setting. Third, the additional modalities enrich primarily the item representations and do not provide comparably rich information about the users.

The ranking formulation also treats every observed interaction as a positive example regardless of its rating. Consequently, one-star and five-star interactions contribute equally to the BPR objective. Moreover, the evaluation ranks each positive item against 99 sampled negatives rather than the complete catalog. Although all models use the same candidates, full-catalog evaluation is required to confirm the findings under a more exhaustive protocol. Finally, the VLM experiment shows that generated visual descriptions are domain-dependent and cannot systematically replace CNN features.

\section{Conclusion}
This paper introduced MPIUR, a graph-based recommendation framework that jointly exploits user reviews, textual item descriptions, and item images. It combines a review-enriched Collaborative Graph, an aspect-level Topic Graph, and an attention-based Modal Encoder. Its objectives align collaborative and semantic representations while giving stronger alignment weight to items with limited interaction histories.

On the Amazon Reviews 2023 Musical Instruments and Baby categories, MPIUR obtains the best global nDCG@10 and Recall@10 among the evaluated methods (Table~\ref{tab:overall}). Its strongest and most consistent performances occur for sparse items. We interpret this pattern as evidence that semantic content is most useful as a complement when collaborative evidence is limited. The ablation results (Table~\ref{tab:ablation}) further suggest that the three modalities provide complementary information: reviews yield the largest individual gain, particularly for sparse items, while image and text features provide smaller gains across all groups. The larger sparse-item degradation after removing the Modality Gap Loss is also consistent with the intended popularity-aware alignment mechanism. We also observe that the removal of modality gap loss or fusion loss affects the performance worse than the removal of image \& text features. It shows that extra input modalities do not necessarily improve performance if they are not ``aligned'' to each other. Unintuitively, the removal of review features degrade the performance on sparse items worse than on popular items, even though we expect that the prediction for the sparse items rely more on the image and text features. A possible reason is that for sparse items, each review strongly affects the prediction, thus their removal affects the performance badly.

The results on popular items require a more cautious interpretation. BM3 remains better on three of the four popular-item metrics, which may indicate that review information becomes redundant or noisy when an item already has a rich interaction history. This explanation remains a hypothesis because the present experiments do not directly measure review redundancy or quality by popularity. Similarly, the VLM experiment (Table~\ref{tab:vlm}) shows that generated visual descriptions are not uniformly superior to CNN features: Qwen2.5-VL improves the performance on Musical Instruments, whereas ResNet50 CNN remains better on Baby. Finally, the HADSF scores (Table~\ref{tab:hadsf-quality}) suggest high opinion fidelity and low aspect drift, while the Topic Graph error analysis (Section~\ref{sec:review_ext_tg_eval}) suggests that the model identifies aspects more reliably than exact opinions. Lastly, the fact that the Topic Graph scores are considerably higher than the random-ranking baseline (Table~\ref{tab:triplet-ranking}) suggests that the graph captures a non-random retrieval signal despite the much larger candidate pool.

The present study remains limited to reviewed interactions, sampled ranking evaluation, and items observed during training. Future work will reintegrate interactions without reviews, incorporate rating signals, enrich user-side representations, combine CNN and VLM visual features, and extend the framework to complete item cold-start.

\begin{acks}
The authors thank the academic supervisors at SUP'COM for their valuable feedbacks.
\end{acks}

\bibliographystyle{ACM-Reference-Format}
\bibliography{sample-base}

@inproceedings{he2020lightgcn,
  title={LightGCN: Simplifying and Powering Graph Convolution Network for Recommendation},
  author={He, Xiangnan and Deng, Kuan and Wang, Xiang and Li, Yan and Zhang, Yongdong and Wang, Meng},
  booktitle={Proceedings of the 43rd International ACM SIGIR Conference on Research and Development in Information Retrieval},
  pages={639--648},
  year={2020}
}

@inproceedings{shuai2023tgnn,
  title={Topic-enhanced Graph Neural Networks for Extraction-based Explainable Recommendation},
  author={Shuai, Jie and Wu, Le and Zhang, Kun and Sun, Peijie and Hong, Richang and Wang, Meng},
  booktitle={Proceedings of the 46th International ACM SIGIR Conference on Research and Development in Information Retrieval},
  pages={1--11},
  year={2023},
  doi={10.1145/3539618.3591776}
}

@inproceedings{zhou2023bm3,
  title={Bootstrap Latent Representations for Multi-modal Recommendation},
  author={Zhou, Xin and Zhou, Hongyu and Liu, Yong and Zeng, Zhiwei and Miao, Chunyan and Wang, Pengwei and You, Yuan and Jiang, Feijun},
  booktitle={Proceedings of the ACM Web Conference 2023},
  pages={845--854},
  year={2023},
  doi={10.1145/3543507.3583251}
}

@inproceedings{vbpr,
  title={VBPR: Visual Bayesian Personalized Ranking from Implicit Feedback},
  author={He, Ruining and McAuley, Julian},
  booktitle={Proceedings of the AAAI Conference on Artificial Intelligence},
  volume={30},
  year={2016}
}

@article{cfmm,
  title={Cross-fusion activates deep modal integration for multimedia recommendation},
  author={Zhang, Chong and Zhang, ZhiCai},
  journal={PLOS ONE},
  volume={20},
  number={7},
  pages={e0327663},
  year={2025},
  doi={10.1371/journal.pone.0327663}
}

@article{koren2009matrix,
  title={Matrix factorization techniques for recommender systems},
  author={Koren, Yehuda and Bell, Robert and Volinsky, Chris},
  journal={Computer},
  volume={42},
  number={8},
  pages={30--37},
  year={2009},
  publisher={IEEE}
}

@inproceedings{wei2019mmgcn,
  title={MMGCN: Multi-modal graph convolution network for personalized recommendation of micro-video},
  author={Wei, Yinwei and Wang, Xiang and Nie, Liqiang and He, Xiangnan and Hong, Richang and Chua, Tat-Seng},
  booktitle={Proceedings of the 27th ACM international conference on multimedia},
  pages={1437--1445},
  year={2019}
}

@article{wang2021dualgnn,
  title={Dualgnn: Dual graph neural network for multimedia recommendation},
  author={Wang, Qifan and Wei, Yinwei and Yin, Jianhua and Wu, Jianlong and Song, Xuemeng and Nie, Liqiang},
  journal={IEEE Transactions on Multimedia},
  volume={25},
  pages={1074--1084},
  year={2021},
  publisher={IEEE}
}

@inproceedings{kompan2010content,
  title={Content-based news recommendation},
  author={Kompan, Michal and Bielikov{\'a}, M{\'a}ria},
  booktitle={International Conference on Electronic Commerce and Web Technologies},
  pages={61--72},
  year={2010},
  organization={Springer}
}

@article{sohn2013contents,
  title={Contents recommendation method using social network analysis},
  author={Sohn, Jong-Soo and Bae, Un-Bong and Chung, In-Jeong},
  journal={Wireless Personal Communications},
  volume={73},
  number={4},
  pages={1529--1546},
  year={2013},
  publisher={Springer}
}

@inproceedings{deepconn,
  title={Joint Deep Modeling of Users and Items Using Reviews for Recommendation},
  author={Zheng, Lei and Noroozi, Vahid and Yu, Philip S.},
  booktitle={Proceedings of the Tenth ACM International Conference on Web Search and Data Mining},
  pages={425--434},
  year={2017}
}

@inproceedings{narre,
  title={Neural Attentional Rating Regression with Review-level Explanations},
  author={Chen, Chong and Zhang, Min and Liu, Yiqun and Ma, Shaoping},
  booktitle={Proceedings of the 2018 World Wide Web Conference},
  pages={1583--1592},
  year={2018}
}

@article{zhang2022latent,
  title={Latent structure mining with contrastive modality fusion for multimedia recommendation},
  author={Zhang, Jinghao and Zhu, Yanqiao and Liu, Qiang and Zhang, Mengqi and Wu, Shu and Wang, Liang},
  journal={IEEE Transactions on Knowledge and Data Engineering},
  volume={35},
  number={9},
  pages={9154--9167},
  year={2022},
  publisher={IEEE}
}

@inproceedings{hu2025modality,
  title={Modality-independent graph neural networks with global transformers for multimodal recommendation},
  author={Hu, Jun and Hooi, Bryan and He, Bingsheng and Wei, Yinwei},
  booktitle={Proceedings of the AAAI Conference on Artificial Intelligence},
  volume={39},
  number={11},
  pages={11790--11798},
  year={2025}
}

@article{li2024graph,
  title={Graph neural networks with deep mutual learning for designing multi-modal recommendation systems},
  author={Li, Jianing and Yang, Chaoqun and Ye, Guanhua and Nguyen, Quoc Viet Hung},
  journal={Information Sciences},
  volume={654},
  pages={119815},
  year={2024},
  publisher={Elsevier}
}

@inproceedings{lin2023printf,
  title={printf: Preference Modeling based on user reviews with item images and textual information via graph learning},
  author={Lin, Hao-Lun and Jiang, Jyun-Yu and Juan, Ming-Hao and Cheng, Pu-Jen},
  booktitle={Proceedings of the 32nd ACM International Conference on Information and Knowledge Management},
  pages={1431--1440},
  year={2023}
}

@misc{lird,
  title={Learning Item Representations Directly from Multimodal Features for Effective Recommendation},
  author={Zhou, Xin and Zhang, Xiaoxiong and Niyato, Dusit and Shen, Zhiqi},
  year={2025},
  eprint={2505.04960},
  archivePrefix={arXiv},
  primaryClass={cs.IR},
  doi={10.48550/arXiv.2505.04960}
}

@misc{gatesid,
  title={GateSID: Adaptive Gating for Semantic-Collaborative Alignment in Cold-Start Recommendation},
  author={Zhu, Hai and Yu, Yantao and Shen, Lei and Wang, Bing and Zeng, Xiaoyi},
  year={2026},
  eprint={2603.22916},
  archivePrefix={arXiv},
  primaryClass={cs.IR},
  doi={10.48550/arXiv.2603.22916}
}

@inproceedings{HADSF,
  title={HADSF: Aspect Aware Semantic Control for Explainable Recommendation},
  author={Nie, Zheng and Sun, Peijie},
  booktitle={Proceedings of the Nineteenth ACM International Conference on Web Search and Data Mining},
  pages={1--11},
  year={2026},
  doi={10.1145/3773966.3778013}
}

@inproceedings{ye2025mllmmsr,
  title={Harnessing Multimodal Large Language Models for Multimodal Sequential Recommendation},
  author={Ye, Yuyang and Zheng, Zhi and Shen, Yishan and Wang, Tianshu and Zhang, Hengruo and Zhu, Peijun and Xiong, Hui},
  booktitle={Proceedings of the AAAI Conference on Artificial Intelligence},
  volume={39},
  number={12},
  year={2025},
  doi={10.1609/aaai.v39i12.33426}
}

@inproceedings{rendle2009bpr,
  title={BPR: Bayesian Personalized Ranking from Implicit Feedback},
  author={Rendle, Steffen and Freudenthaler, Christoph and Gantner, Zeno and Schmidt-Thieme, Lars},
  booktitle={Proceedings of the Twenty-Fifth Conference on Uncertainty in Artificial Intelligence},
  pages={452--461},
  year={2009}
}

@misc{qwen2025vl,
  title={Qwen2.5-VL Technical Report},
  author={{Qwen Team}},
  year={2025},
  eprint={2502.13923},
  archivePrefix={arXiv},
  primaryClass={cs.CV}
}

@misc{dubey2024llama,
  title={The Llama 3 Herd of Models},
  author={Dubey, Abhimanyu and others},
  year={2024},
  eprint={2407.21783},
  archivePrefix={arXiv},
  primaryClass={cs.AI}
}

@misc{hou2024amazon,
  title={Bridging Language and Items for Retrieval and Recommendation},
  author={Hou, Yupeng and Li, Jiacheng and He, Zhankui and Yan, An and Chen, Xiusi and McAuley, Julian},
  year={2024},
  eprint={2403.03952},
  archivePrefix={arXiv},
  primaryClass={cs.IR}
}

@inproceedings{liu2023llava,
  author      = {Liu, Haotian and Li, Chunyuan and Wu, Qingyang and Lee, Yong Jae},
  title       = {Visual Instruction Tuning},
  booktitle   = {NeurIPS},
  year        = {2023}
}

\section*{GenAI-Usage Disclosure}   
Generative AI assistants (Claude by Anthropic as well as Mistral by Mistral AI) was used in a limited, non-authorial capacity during the preparation of this work: for minor code corrections and debugging assistance, for LaTeX formatting and compilation troubleshooting, and for language/wording suggestions during manuscript editing. The research design, model architecture, experiments, result analysis, and scientific claims presented in this paper are entirely the authors' own work; no AI-generated text, code, or results were incorporated without review and validation by the authors.

\end{document}